\pdfoutput=1

\documentclass[11pt]{article}

\usepackage[final]{acl}

\usepackage{times}
\usepackage{latexsym}

\usepackage[T1]{fontenc}

\usepackage[utf8]{inputenc}

\usepackage{microtype}

\usepackage{inconsolata}

\usepackage{graphicx}

\usepackage{amsfonts}
\usepackage{amsmath}
\usepackage{xcolor}
\usepackage{multirow}
\usepackage{booktabs}
\usepackage{algorithm,algorithmic}
\usepackage{amssymb}
\usepackage{bbding}
\usepackage{listings}
\usepackage[most]{tcolorbox}
\usepackage{subcaption}
\usepackage{fontawesome}

\title{ProBench: Benchmarking Large Language Models in Competitive Programming}

\author{%
\\ \textbf{Lei Yang}\quad 
  \textbf{Renren Jin} \quad 
  \textbf{Ling Shi} \quad 
  \textbf{Jianxiang Peng} \quad \\
  \textbf{Yue Chen} \quad 
  \textbf{Deyi Xiong$^{*}$} \quad 
  \vspace{0.5em} \\
  Tianjin University \\
  \{yanglei\_9, dyxiong\}@tju.edu.cn \\
    \faGithub\ \href{https://github.com/YL-9/probench}{https://github.com/YL-9/probench}
  }

\begin{document}
\maketitle

\begin{abstract}
With reasoning language models such as OpenAI-o3 and DeepSeek-R1 emerging, large language models (LLMs) have entered a new phase of development. However, existing benchmarks for coding evaluation are gradually inadequate to assess the capability of advanced LLMs in code reasoning. To bridge the gap for high-level code reasoning assessment, we propose ProBench to benchmark LLMs in competitive programming, drawing inspiration from the International Collegiate Programming Contest. ProBench collects a comprehensive set of competitive programming problems from Codeforces, Luogu, and Nowcoder platforms during the period from July to December 2024, obtaining real test results through online submissions to ensure the fairness and accuracy of the evaluation. We establish a unified problem attribute system, including difficulty grading and algorithm tagging. With carefully collected and annotated data in ProBench, we systematically assess 9 latest LLMs in competitive programming across multiple dimensions, including thought chain analysis, error type diagnosis, and reasoning depth evaluation. Experimental results show that QwQ-32B-Preview achieves the best score of 20.93 followed by DeepSeek-V3 with a score of 16.38, suggesting that models trained with specialized reasoning tasks significantly outperform general-purpose models (even larger than reasoning-oriented models) in programming. Further analysis also reveals key areas for programming capability enhancement, e.g., algorithm adaptability and reasoning sufficiency, providing important insights for the future development of reasoning models.
\end{abstract}

\begin{figure*}
    \centering
    \includegraphics[width=13cm]{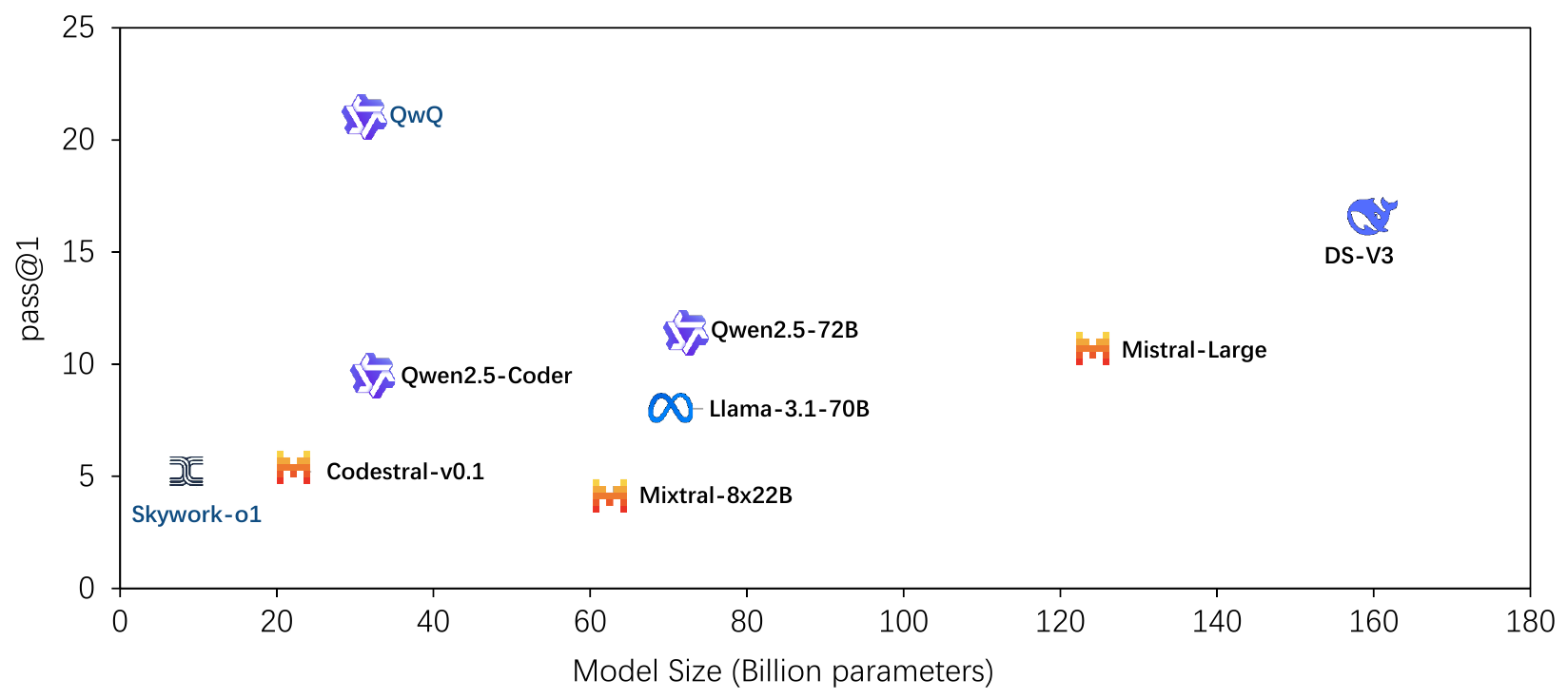}
    \caption{The pass@1 results of all evaluated models on ProBench. Model names in blue are reasoning models while the others are non-reasoning models.}
    \label{fig:main}
\end{figure*}

\section{Introduction}

As OpenAI-o3 \cite{o3} and DeepSeek-R1 \cite{deepseek-r1} emerge, reasoning language models have achieved unprecedented progress in advanced reasoning. These models have not only demonstrated breakthroughs in traditional STEM-related benchmarks such as mathematics \cite{hendrycks2021measuringmath, cobbe2021training, rein2023gpqa} and physics \cite{welbl2017crowdsourcing}, but have also gained significant improvements in programming \cite{hendrycks2021measuringapps}, exhibiting impressive reasoning and coding competence. However, existing code evaluation benchmarks are usually not adequate to assess advanced LLMs in challenging programming, especially competitive programming.

Specifically, competitive programming requires participants to analyze problems, select appropriate data structures and algorithms, and implement code that satisfies rigorous time-space efficiency constraints and boundary conditions. Codes corresponding to these problems must pass extensive targeted test cases under predefined evaluation criteria. However, these test cases are typically accessible only to competition organizers rather than being publicly available, with participants merely permitted to submit code for correctness verification. In contrast to this, existing code benchmarks usually suffer from the lack of robust test suites to comprehensively validate code robustness, thereby compromising evaluation fairness (Challenge 1). Additionally, most current evaluation efforts remain superficial, focusing solely on measuring code submission pass rates without conducting thorough and systematic analyses of model capabilities (Challenge 2).

To address these challenges, inspired by the International Collegiate Programming Contest (ICPC), we propose ProBench, which is designed to comprehensively, fairly, and thoroughly analyze the reasoning capability of LLMs in competitive programming. We collect all competition problems from July to December 2024 on renowned programming platforms including Codeforces,\footnote{\url{https://codeforces.com}} Luogu,\footnote{\url{https://www.luogu.com.cn}} and Nowcoder.\footnote{\url{https://ac.nowcoder.com}} Notably, problem descriptions from Codeforces are in English, while the latter two platforms primarily in Chinese. In addition to problem descriptions, we gather extensive problem metadata such as difficulty levels and algorithmic tags for subsequent in-depth analysis.

We then provide LLMs with all information accessible to human participants during programming competitions lanuched by these programming platforms to generate solution ideas and codes. Unlike previous benchmarks, we propose to submit the generated code solutions to the original competition platform. This enables the utilization of its comprehensive test cases to rigorously assess code correctness, while simultaneously acquiring feedback results that can be systematically employed for subsequent analytical processes. This submission-based evaluation ensures code robustness through rigorous testing under standardized environments, demonstrating superior fairness and accuracy compared to offline evaluation approaches adopted by other benchmarks, hence addressing Challenge 1.

With created ProBench and its evaluation strategy, we systematically evaluate 7 prevalent non-reasoning models and 2 reasoning models. As illustrated in Figure \ref{fig:main}, experimental results demonstrate that reasoning models exhibit significant advantages in code reasoning over non-reasoning models. Notably, QwQ-32B-Preview achieves the highest score of 20.93 points.

To address Challenge 2, we perform a multidimensional investigation combining chain-of-thought analysis and code evaluation to thoroughly examine the code reasoning capability of LLMs. Through comprehensive data analysis, we systematically reveal issues and performance discrepancies among different models during reasoning processes.

The main contributions of our work can be summarized into three aspects as follows.

\begin{itemize}

    \item We propose ProBench that assesses LLMs in competitive programming, satisfying the evaluation demand for emerging reasoning language models. 

    \item We pioneer an online submission mechanism that ensures fairness and validity in code robustness assessment.

    \item Through extensive experiments, we comprehensively analyze patterns of code reasoning in LLMs, providing insights for future reasoning enhancement.

\end{itemize}

\begin{table*}[t]
    \centering
    \small
    \begin{tabular}{ccccccccc}
    \toprule
        \multirow{2}{*}{Benchmark} & Release & \multirow{2}{*}{Difficulty} & Num of & Description & \multirow{2}{*}{Update} & Online Code & Multi-Site & In-depth \\
        & Date &  & Probs & Language &  & Evaluation & Sources & Analysis \\ 
        \midrule
        APPS & 2021/05 & $\bigstar \bigstar$ & 10000 & EN & \XSolidBrush & \XSolidBrush & \Checkmark & \XSolidBrush \\ 
        CodeContests & 2022/03 & $\bigstar \bigstar \bigstar \bigstar$ & 165 & EN & \XSolidBrush & \XSolidBrush & \Checkmark & \XSolidBrush \\ 
        xCodeEval & 2023/03 & $\bigstar \bigstar$ & 952 & EN & \XSolidBrush & \XSolidBrush & \XSolidBrush & \XSolidBrush \\ 
        LeetCode-Hard & 2023/03 & $\bigstar$ & 40 & EN & \XSolidBrush & \XSolidBrush & \XSolidBrush & \XSolidBrush \\ 
        TACO & 2023/12 & $\bigstar \bigstar$ & 1000 & EN & \XSolidBrush & \XSolidBrush & \Checkmark & \XSolidBrush \\ 
        LiveCodeBench & 2024/03 & $\bigstar$ & 511 & EN & \Checkmark & \XSolidBrush & \Checkmark & \XSolidBrush \\ 
        USACO & 2024/04 & $\bigstar \bigstar$ & 307 & EN & \Checkmark & \XSolidBrush & \XSolidBrush & \XSolidBrush \\ 
        CodeElo & 2025/01 & $\bigstar \bigstar \bigstar$ & 387 & EN & \Checkmark & \Checkmark & \XSolidBrush & \XSolidBrush \\ 
        \midrule
        ProBench & 2025/02 & $\bigstar \bigstar \bigstar \bigstar \bigstar$ & 790 & EN, CN & \Checkmark & \Checkmark & \Checkmark & \Checkmark \\ 
        \bottomrule
    \end{tabular}
    \caption{Comparison of the ProBench benchmark against previous related benchmarks.}
    \label{tab:benchmarks}
\end{table*}

\section{Related Work}

\textbf{Code Language Models and Reasoning Language Models.} Recent years have witnessed emerging LLMs specifically tailored for code-related tasks, such as AlphaCode \cite{li2022competition}, CodeLLaMa \cite{roziere2023code}, StarCoder \cite{li2023starcoder} and CodeGeeX \cite{zheng2023codegeex}. Trained on extensive open-source code repositories, these models demonstrate exceptional performance in code generation benchmarks \cite{chen2021evaluating, austin2021program, ni2024l2ceval} and code search tasks \cite{lu2021codexglue, huang2021cosqa, khan2023xcodeeval}. Despite significantly enhancing code generation, they exhibit limitations in reasoning capabilities, struggling to independently tackle tasks requiring deep logical reasoning, such as competition-level programming and complex engineering requirements. Recent advancements in reasoning language models, such as OpenAI-o3 \cite{o3}, DeepSeek-R1 \cite{deepseek-r1}, and QwQ-32B-Preview \cite{qwq}, trained via chain-of-thought prompting and reinforcement learning techniques, have demonstrated human-competitive proficiency in handling intricate programming challenges. In domains demanding reasoning capabilities comparable to mathematical competitions \cite{hendrycks2021measuringmath, cobbe2021training, rein2023gpqa}, these models have even surpassed the majority of human participants, presenting unprecedented challenges to traditional evaluation benchmarks in programming competitions.

\textbf{Code Generation Benchmarks.} Numerous studies have focused on evaluating code generation capabilities of models through established benchmarks such as HumanEval \cite{chen2021evaluating} and MBPP \cite{austin2021program}. These benchmarks primarily assess model performance in generating standalone function implementations by designing fine-grained function-level coding tasks (e.g., implementing specific algorithms or data operations) and validating code correctness via automated test cases, providing a robust foundation for capability evaluation. However, such testing scenarios often remain overly idealized, emphasizing isolated function implementations while neglecting the complexity of code in real-world scenarios.

\textbf{Competitive Programming Benchmarks.} In addition, there exist specialized benchmarks designed for programming competitions such as APPS \cite{hendrycks2021measuringapps}, CodeContests \cite{li2022competition}, xCodeEval \cite{khan2023xcodeeval}, LeetCode-Hard \cite{shinn2024reflexion}, TACO \cite{li2023taco}, LiveCodeBench \cite{jain2024livecodebench} and USACO \cite{shi2024can}. These benchmarks not only substantially exceed the aforementioned code generation benchmarks in problem complexity but also impose significantly higher demands on models' reasoning capabilities. However, these benchmarks predominantly adopt offline evaluation methods, replacing the original platform’s robust test cases with locally curated test cases, which frequently leads to significant false positive results. While CodeElo \cite{quan2025codeelo}, a contemporaneously released benchmark, employs an online submission strategy similar to ours, its analysis of model reasoning capabilities remains incomplete, failing to comprehensively capture model performance in complex reasoning scenarios. In contrast, our benchmark provides an in-depth analysis of the reasoning capabilities of LLMs through comprehensive data. Table \ref{tab:benchmarks} provides a comprehensive comparison of our benchmark with existing benchmarks across multiple dimensions.

\begin{table}[t]
    \centering
    \small
    \begin{tabular}{c c c c c c}
    \toprule
    \multirow{2}{*}{Website} & \multirow{2}{*}{Lang} & Problem  & \multicolumn{3}{c}{Difficulty} \\
     & & Count & easy & medium & hard \\
    \midrule
    Codeforces & EN & 446 & 173 & 139 & 110 \\
    Luogu & CN & 63  & 30  & 23  & 10  \\
    Nowcoder & CN & 281 & 102 & 40  & 12  \\
    \midrule
    Total & -  & 790 & 305 & 202 & 132 \\
    \bottomrule
    \end{tabular}
    \caption{Statistics of ProBench.}
    \label{tab:problem_stats}
\end{table}

\section{ProBench}

We elaborate ProBench with a detailed account of its data collection, attribute fusion, and online submission process. Statistics of ProBench are presented in Table \ref{tab:problem_stats}.

\subsection{Data Collection}

To enhance the diversity of problem sets and mitigate data contamination, we collect all competition problems from three prominent programming contest platforms (Codeforces, Luogu, and Nowcoder) spanning the period from July to December 2024. Notably, problem descriptions on Codeforces are exclusively presented in English, while those on Luogu and Nowcoder are provided in Chinese, establishing an effective framework for evaluating model capabilities in multilingual reasoning. Furthermore, we preserve comprehensive problem attributes, including difficulty levels, algorithm tags, and creation timestamps, to facilitate in-depth analysis of model performance. The problem descriptions (an example is presented in Appendix \ref{sec:prob_d}) maintain strict consistency with the information accessible to human participants during actual competitions.

\subsection{Attribute Integration}

Problem attributes vary significantly across different programming platforms. To facilitate systematic analysis, we perform unified integration of problem difficulty levels and algorithm tags from heterogeneous sources to establish consistent representations.

\textbf{Problem Difficulty.} We systematically consolidate and standardize difficulty descriptions from multiple platforms into three unified tiers : Easy, Medium, and Hard, corresponding to ICPC award criteria. For instance, Codeforces employs integer values within [800, 3500] to denote problem difficulty. To align with our grading system, we categorize [800, 1500] as Easy, (1500, 2400] as Medium, and (2400, 3500] as Hard. A comprehensive cross-platform difficulty mapping is provided in Appendix \ref{sec:integration}.

\textbf{Algorithm Tags.} Solving programming problems requires not only profound thinking and reasoning skills but also mastery of relevant data structures and algorithms, collectively termed as algorithm tags. Given the heterogeneous taxonomy across platforms, we normalize these tags into seven knowledge categories: Basic, Search, String, Dynamic Programming (DP), Data Structures (DS), Graph, and Mathematics (Math). This categorization scheme considers both the cognitive requirements of different algorithms and enables precise analysis of models' logical reasoning capabilities. Specific classification criteria and implementation details are documented in Appendix \ref{sec:integration}.

\subsection{Online Submission}

To rigorously verify code robustness, it is essential to design test cases targeting exceptional scenarios in addition to regular test cases. In testing environments involving large data volumes or special conditions (e.g., chrysanthemum graphs, tree degeneration into chains), the code must employ optimized data structures and algorithms to satisfy predefined temporal and spatial constraints. However, generating such specialized test cases typically requires substantial effort from problem setters and is often confined within the internal environments of programming competition platforms rather than being publicly disclosed.

To address this challenge, we adopt a online submission strategy that directly submits model-generated code to the original competition platform's online evaluation system. This approach leverages the platform's proprietary test cases to comprehensively assess code robustness, thereby eliminating potential false positive outcomes. Upon completion of the evaluation, we systematically collect and archive detailed results, including failure causes (e.g., compilation errors, incorrect answers), obtain scores and runtime resource consumption (time and memory usage) for subsequent experimental analysis.

\begin{table*}[t]
    \centering
    \small
    \begin{tabular}{c c c c c c c c c c c c}
        \toprule
        \multirow{2}{*}{Model} & \multirow{2}{*}{Size} & \multirow{2}{*}{R} & \multicolumn{4}{c}{pass@} & \multicolumn{3}{c}{pass@1 for 3 levels} & \multicolumn{2}{c}{Lang} \\
              &            &                 & 1           & 2           & 4           & 8           & easy    & medium  & hard    & EN  & CN \\
        \midrule
        QwQ              & 32B      & \Checkmark & 20.93 & 26.43 & 31.35 & 36.08 & 40.66 & 2.62  & 0.00 & 18.93 & 23.80 \\
        DS-V3                  & 37/671B  & \XSolidBrush & 16.38 & 20.24 & 23.67 & 26.58 & 31.76 & 0.80  & 0.00 & 12.39 & 21.55 \\
        Qwen2.5-72B         & 72B      & \XSolidBrush & 11.50 & 14.39 & 16.97 & 19.24 & 23.24 & 0.37  & 0.00 & 8.66  & 15.19 \\
        Mistral-Large  & 123B     & \XSolidBrush & 10.54 & 13.87 & 17.26 & 20.89 & 20.82 & 0.37  & 0.00 & 8.07  & 13.74 \\
        Qwen2.5-Coder   & 32B      & \XSolidBrush & 9.48  & 12.73 & 15.80 & 18.48 & 17.91 & 0.56  & 0.00 & 5.41  & 14.75 \\
        Llama-3.1-70B       & 70B      & \XSolidBrush & 7.99  & 10.15 & 12.50 & 15.06 & 16.23 & 0.06  & 0.00 & 5.80  & 10.83 \\
        Codestral-v0.1           & 22B      & \XSolidBrush & 5.08  & 7.08  & 9.36  & 11.65 & 10.70 & 0.00  & 0.00 & 3.59  & 7.01  \\
        Skywork-o1   & 8B       & \Checkmark & 5.06  & 6.80  & 8.48  & 10.13 & 10.53 & 0.00  & 0.00 & 3.53  & 7.05  \\
        Mixtral-8x22B    & 22/176B  & \XSolidBrush & 4.27  & 5.85  & 7.49  & 9.11  & 8.61  & 0.00  & 0.00 & 2.83  & 6.14  \\
        \bottomrule
    \end{tabular}
    \caption{Evaluation results of the examined models on ProBench, sorted by pass@1. For detailed model information, please refer to the Table \ref{tab:models}. $``\text{R}"$ indicates reasoning-oriented models.}
    \label{tab:model_performance}
\end{table*}

\section{Experiments}

We conducted a series of experiments to evaluate recent advanced LLMs (including reasoning-oriented models) on ProBench.

\subsection{Settings}

\textbf{Models.} We evaluated 9 prominent open-source LLMs spanning instruction-tuned, code-specialized, and reasoning-optimized models. During preliminary analysis, we observed that non-reasoning models with fewer parameters exhibit limited reasoning capabilities, often failing to complete even elementary tasks. To maintain evaluation efficiency and prevent the inclusion of underperforming models in the competition platform, we consequently excluded non-reasoning models with below 14B parameters from our formal assessment. The evaluated models are categorized as follows (see Appendix \ref{sec:model_d} for detailed specifications):

\begin{enumerate}

    \item Instruction-tuned Models: DeepSeek-V3, Mistral-Large-Instruct-2411, Mixtral-8x22B-Instruct-v0.1, Qwen2.5-72B-Instruct, Llama-3.1-70B-Instruct.

    \item Code-specialized Models: Codestral-22B-v0.1, Qwen2.5-Coder-32B-Instruct.

    \item Reasoning-optimized Models: QwQ-32B-Preview, Skywork-o1-Open-Llama-3.1-8B.

\end{enumerate}

\textbf{Evaluation Metrics.} Unlike other benchmarks \cite{hendrycks2021measuringapps, shi2024can, jain2024livecodebench} that predominantly use Python as the target language for model-generated code, we adopted C++ as the primary evaluation language given its prevalence in competitive programming. Additionally, we analyzed two other widely-used languages, Java and Python, in Section \ref{sec:c_lang}. To accommodate different problem description languages, we employed both English and Chinese prompts, with detailed prompt templates provided in Appendix \ref{sec:prompt}.

For model evaluation, we primarily utilized the established pass@k metric \cite{kulal2019spoc, chen2021evaluating} by generating 8 candidate solutions per problem. While explicit pass@k results were reported where applicable, all other evaluations defaulted to pass@1. During code generation, we maintained each model's default hyperparameters (including temperature, \text{top\_p}, and \text{top\_k}) while setting \text{max\_tokens} to the model's default context length minus the prompt length.

\subsection{Main Results}

Table \ref{tab:model_performance} presents the main evaluation results of examined models. It can be observed that as the number of samples increases, the accuracy of all models exhibits an upward trend. Notably, the reasoning model QwQ-32B-Preview, with merely 32 billion parameters, achieves a leading pass@1 score of 20.93, surpassing DeepSeek-V3 despite the latter's substantially larger parameter scale. Furthermore, the 8B-sized Skywork-o1-Open-Llama-3.1-8B demonstrates superior performance compared to the 22B/176B-sized Mixtral-8x22B-Instruct-v0.1, while maintaining comparable efficacy with the code-specialized model Codestral-22B-v0.1. These findings suggest that reasoning-oriented training plays a pivotal role in significantly enhancing model capabilities, while also indicating that the foundational performance of base models directly influences the outcomes of their post-training variants. The non-reasoning model DeepSeek-V3 followes closely with a score of 16.38. Additionally, our analysis reveals that Qwen2.5-72B-Instruct marginally outperforms the 123B-sized Mistral-Large-Instruct-2411.

\textbf{Difficulty.} As evidenced by the results presented in Table \ref{tab:model_performance}, model scores exhibit a marked decline with increasing problem difficulty. At the $``\text{easy}"$ difficulty level, examined LLMs demonstrates strong capability with scores ranging from 8.61 to 40.66, whereas their scores approach 0 for both $``\text{medium}"$ and $``\text{hard}"$ difficulty tiers. This observation indicates that our benchmark not only effectively differentiates disparities in models' reasoning capacities but also possesses sufficient rigor to accommodate the evaluation demands of reasoning models for the foreseeable future.

\begin{figure*}
    \centering
    \includegraphics[width=10.9cm]{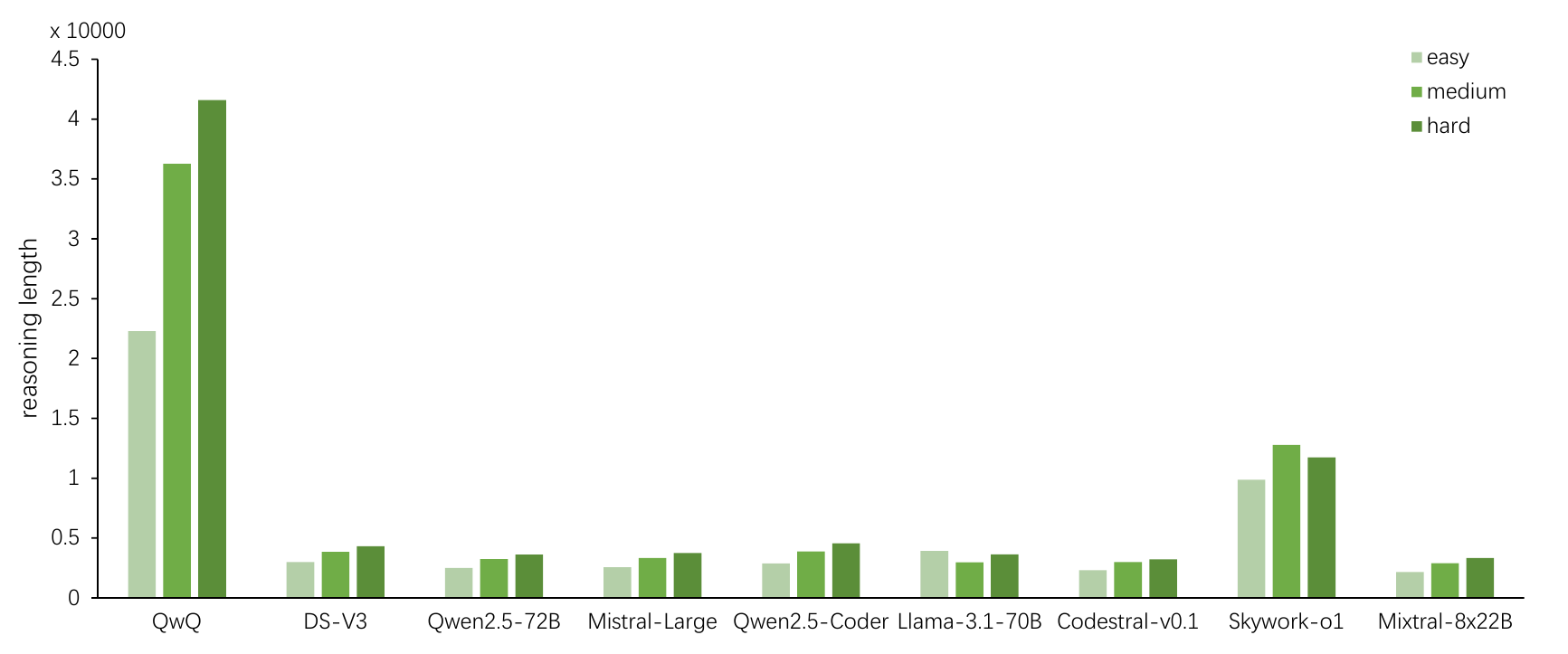}
    \caption{Presents the CoT length, measured in characters, for each model, ranked by inference capability.}
    \label{fig:l_cot}
\end{figure*}

\begin{figure*}
    \centering
    \includegraphics[width=8.9cm]{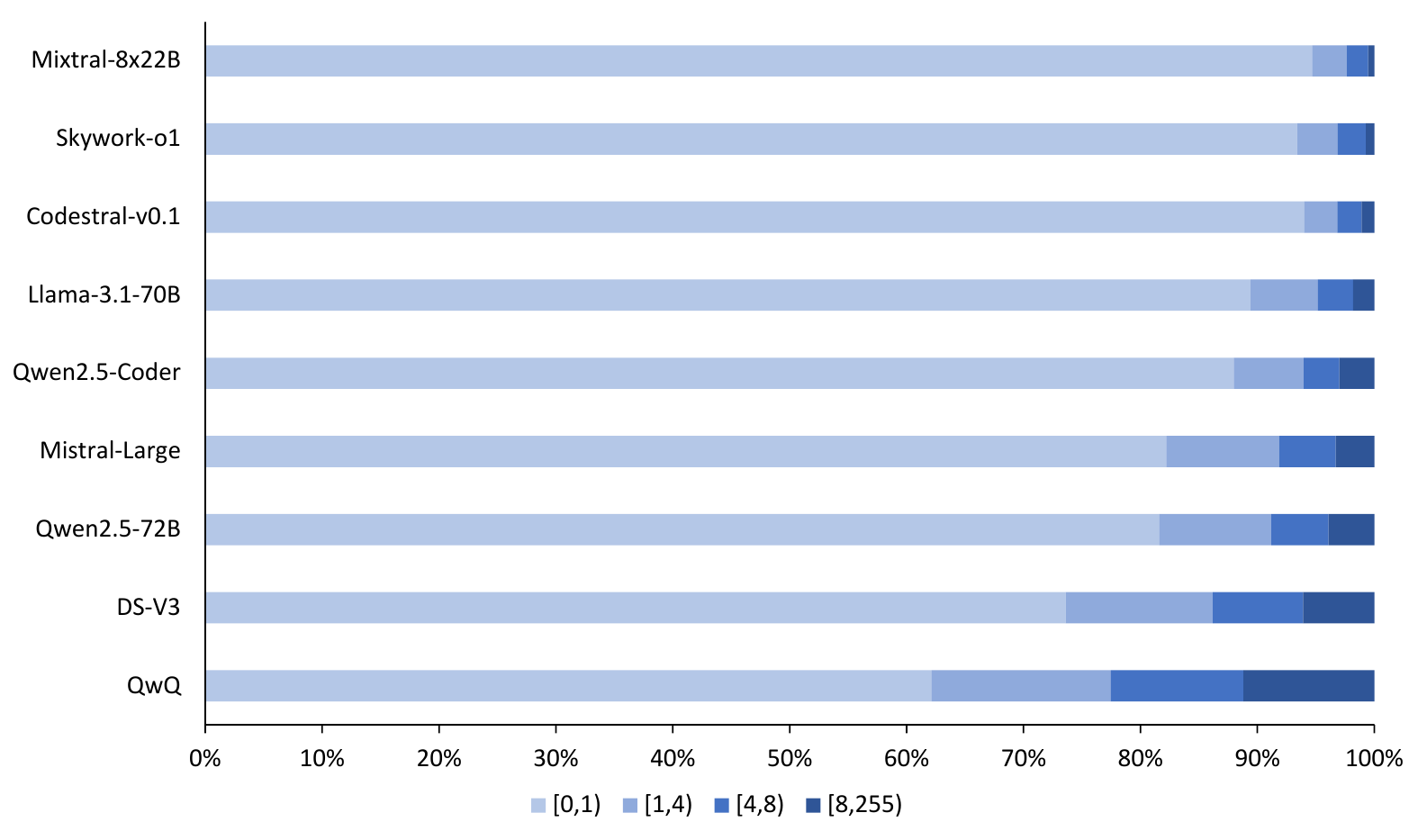}
    \caption{Presents the ratio of the sum of error intervals in the code generated by each model. The interval $[1,4)$ indicates the number of failed code instances within the $[1,4)$ range of test cases.}
    \label{fig:error_p}
\end{figure*}

\textbf{Multilingual Competence.} Further analysis of language-specific performance, after controlling for variations in problem difficulty, reveal problem horizontal comparisons of model performance across Chinese and English contexts. The results demonstrate robust bilingual support across all evaluated models, with no instances of inadequate linguistic adaptation. Concurrently, model scores exhibit a natural degradation pattern that correlates with diminishing reasoning capabilities, maintaining logical consistency in performance trends.


\section{Analysis}

To conduct an in-depth analysis of the reasoning capability of LLMs, we integrated the inherent attributes of programming problems with the outcomes of code evaluations, providing detailed discussions from five distinct perspectives.

\subsection{Length of CoT}

To investigate whether models exhibit under-reasoning or over-reasoning \cite{chen2024not} during inference, we conducted systematic analysis of model-generated responses. Considering that human perception of reasoning length primarily relies on textual character count rather than token quantity, we adopted character length as the metric for measuring chain-of-thought (CoT) \cite{wei2022chain} complexity.

Figure \ref{fig:l_cot} reveals an overall increasing trend in reasoning length with elevated problem difficulty. However, most models demonstrate relatively modest growth magnitudes in reasoning length, approximately 30\% from easy to medium difficulty level, and merely 15\% from medium to hard. This pattern suggests potential under-reasoning phenomena, indicating that these models might engage in insufficient deliberation when confronted with high-intensity reasoning tasks.

Notably, reasoning-oriented models exhibit significantly longer CoT sequences (10k of characters on average) compared to non-reasoning models, even when addressing simple problems. This phenomenon implies potential over-reasoning tendencies in reasoning-oriented models when handling low-difficulty tasks. Nevertheless, despite their generally extended reasoning lengths across various difficulty levels, the incremental growth pattern of reasoning-oriented models remains relatively constrained (only 60\% and 15\%) as problem complexity increases, further corroborating the persistence of under-reasoning issues.



\begin{figure*}
    \centering
    \includegraphics[width=7cm]{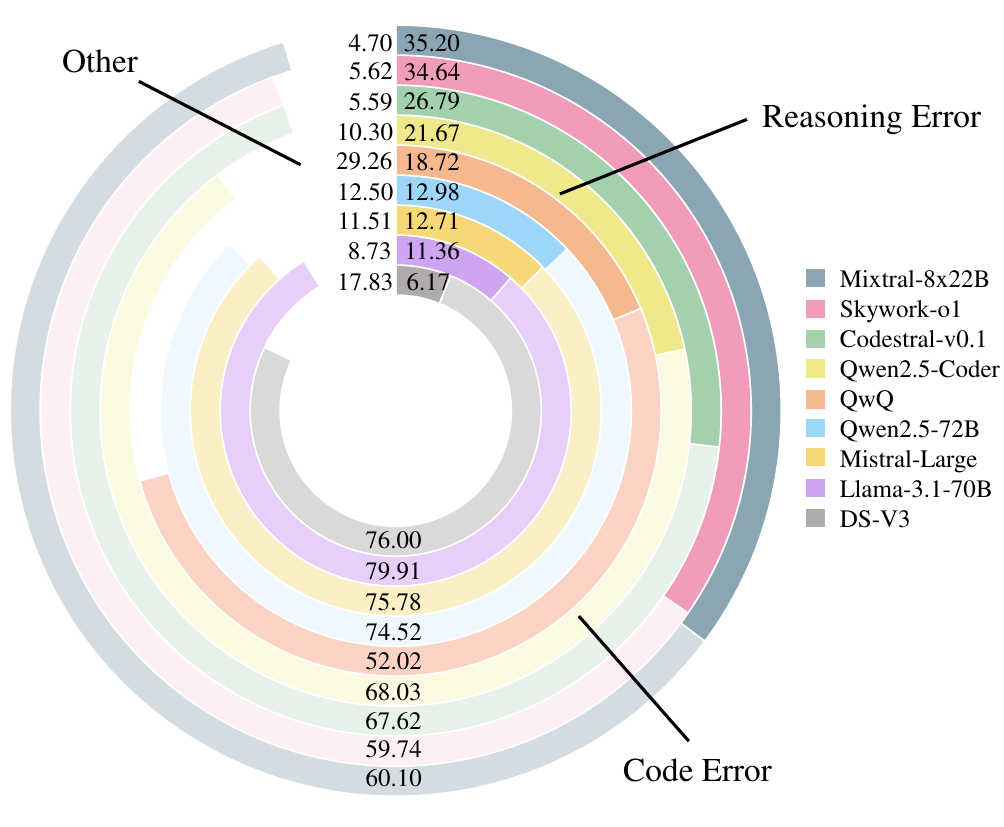}
    \caption{Presents the distribution of error types in the code generated, with the proportion of reasoning errors increasing from the innermost to the outermost layers.}
    \label{fig:code_er}
\end{figure*}

\subsection{Deep Reasoning}

Furthermore, we conducted statistical analysis on the error points of codes submitted to the Codeforces platform, which fail to pass. Generally, the earlier the error points occur, the lower their detection difficulty tends to be, primarily focusing on verifying whether the code can handle input/output operations correctly without considering execution efficiency. Consequently, error points appearing earlier indicate shallower reasoning depth in the models, often limited to superficial logical judgments or even a failure to accurately comprehend problem requirements. Such models cannot select appropriate data structures and algorithms for deeper reasoning to optimize code efficiency.

Figure \ref{fig:error_p} demonstrates that models with stronger reasoning capabilities tend to exhibit error occurrences predominantly at later test positions. For instance, the most advanced reasoning model QwQ-32B-Preview exhibits only 62\% error rate on the initial test case (Case 0), whereas the weaker Mixtral-8x22B-Instruct-v0.1 model surpasses 90\% under equivalent conditions. This observation indicates a strong correlation between the reasoning capability of a model and its ability to conduct deep, multi-step reasoning processes.

\subsection{Fundamental Capabilities}

To thoroughly analyze the error types in model-generated code, we submitted the outputs to the original programming platforms and collected specific failure reasons for rejected solutions. For instance, on Codeforces, error categories include $``\text{WRONG\_ANSWER}"$. This methodology provides direct insights into models' fundamental capabilities in generating executable code and their advanced reasoning competencies.

In this analysis, we evaluated results from Codeforces and Nowcoder platforms, categorizing errors into two primary classes: code errors and reasoning errors. Code errors encompass issues like $``\text{COMPILATION\_ERROR}"$, primarily reflecting syntax or execution problems. Reasoning errors include $``\text{WRONG\_ANSWER}"$, revealing deficiencies in logical reasoning and algorithm optimization (See Appendix \ref{sec:error_c} for details).

As illustrated in Figure \ref{fig:code_er}, the analysis reveals distinct patterns. Models with smaller parameter sizes exhibit higher proportions of code errors, indicating substantial room for improvement in basic code generation capabilities. Notably, even QwQ-32B-Preview, the model with strongest reasoning performance, demonstrates unsatisfactory code error rates (18.72\%). However, as model scale increases, code error proportions gradually decrease while reasoning errors become relatively more prominent. This suggests that increasing model capacity generally strengthens fundamental code generation abilities, yet the persistent prevalence of reasoning errors underscores inherent limitations in solving complex algorithmic problems. The inverse correlation between code error reduction and reasoning error escalation implies that while models achieve better code validity through scaling, their reasoning capabilities remain insufficient for high-difficulty programming challenges, necessitating further advancement in logical deduction and optimization strategies.


\begin{figure*}
    \centering
    \includegraphics[width=8cm]{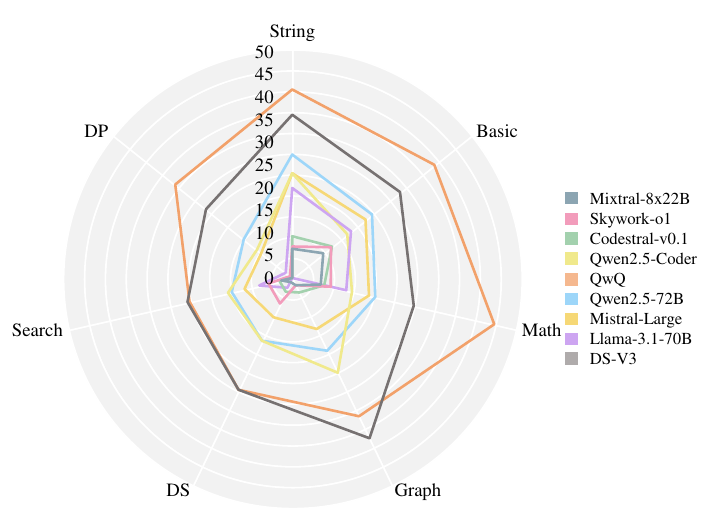}
    \caption{Presents the performance across different data structures and algorithms. As the rotation proceeds clockwise, the difficulty of reasoning gradually increases.}
    \label{fig:alg_tags}
\end{figure*}

\subsection{Algorithmic Tags}

Significant variations exist in the reasoning complexity of data structure and algorithm problems across different difficulty levels. Given the model's inferior performance on problems with medium and hard difficulty level, we exclusively analyzed easy level problems. At the easy level, our comprehensive evaluation established the following reasoning complexity hierarchy: String < Basic < Math < Graph < Data Structures (DS) < Search < Dynamic Programming (DP). See Appendix \ref{sec:rch} for detailed analysis.

Figure \ref{fig:alg_tags} reveals distinct disparities in the model's capability to process different data structure and algorithmic problems. Specifically, models achieve superior performance on problems with lower reasoning intensity. However, their effectiveness progressively diminishes with increasing reasoning demands, indicating persistent limitations in handling problems requiring advanced logical reasoning capabilities even when confronting problems of comparable difficulty levels.

\subsection{Code Language} \label{sec:c_lang}

\begin{figure}
    \centering
    \includegraphics[width=6cm]{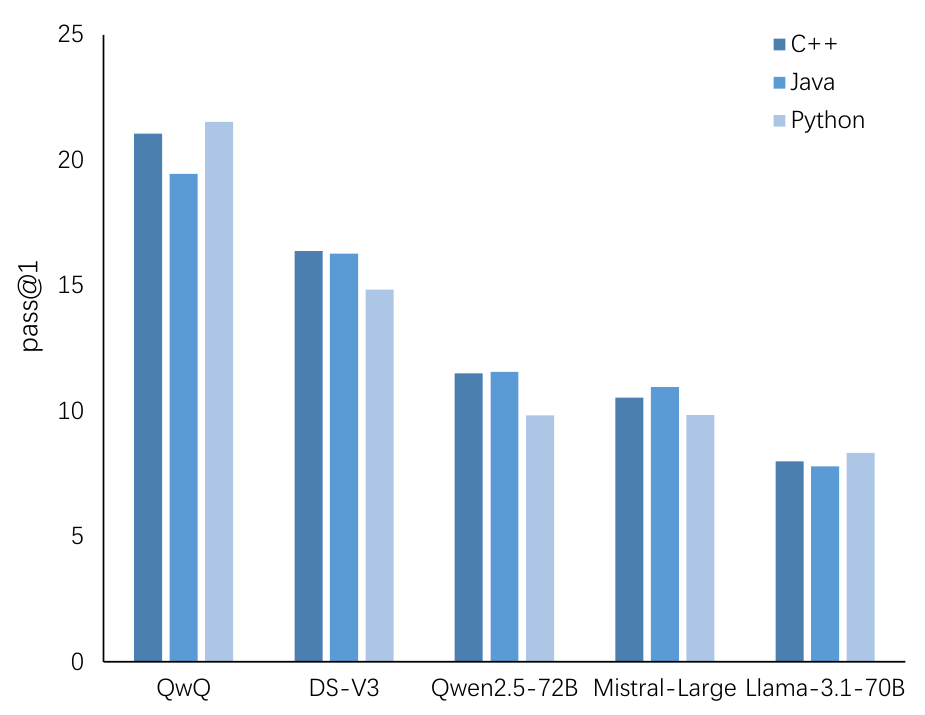}
    \caption{The model's capability to generate different programming languages.}
    \label{fig:code_lang}
\end{figure}

In addition to C++, we selected five models to generate code in Java and Python, commonly used languages among programming contestants, to investigate their code generation capabilities across different programming languages. Figure \ref{fig:code_lang} demonstrates that evaluation results across the three programming languages remain largely consistent, fluctuating within approximately 2\%. This suggests that the reasoning capabilities of the models are largely independent of the programming languages used, with similar learning capacities observed across the three programming languages.

\section{Conclusion}

In this paper, we have presented the ProBench evaluation benchmark, designed to conduct comprehensive, fair, and in-depth analysis of the code reasoning capabilities of LLMs. The benchmark not only enables precise assessment of existing models' reasoning proficiency but also provides substantial evaluation space for the advancement of future reasoning models. ProBench collects substantial problem sets with verified labels from three programming platforms, employing original platform verification through online code submission to effectively eliminate interference from false positive results. In our experimental evaluation of 9 open-source models, the results demonstrate that smaller-scale reasoning models outperform non-reasoning models with significantly larger model sizes. This finding underscores the critical role of reasoning capability in model performance. More importantly, through extensive experimentation, we systematically analyze model reasoning capabilities and propose critical insights aimed at advancing future research and development in reasoning language models.

\section*{Limitation}

ProBench primarily relies on three major programming contest platforms, Codeforces, Luogu, and Nowcoder, for code evaluation, which currently lags behind offline evaluation systems in terms of operational convenience. To enhance the evaluation experience, we will actively explore novel solutions that improve evaluation accessibility while maintaining system robustness in code assessment.

As of current testing phases, the number of evaluated reasoning language models remains limited. This constraint primarily stems from the substantial computational reasoning requirements of DeepSeek-R1 and its distilled variants on our benchmark tasks, coupled with their comparatively slow inference speeds. These factors collectively result in significantly prolonged execution durations, often extending to tens or even hundreds of times longer than those required by non-reasoning models. We plan to expedite the release of comprehensive evaluation results for these models to better demonstrate their performance characteristics.

\section*{Ethical Statement}

We sincerely express our gratitude to the Codeforces, Luogu, and Nowcoder platforms for their exceptional infrastructure, which has provided substantial support for our research. Throughout the experimental process, we strictly adhered to the terms of use of Codeforces,\footnote{\url{https://codeforces.com/terms}} Luogu,\footnote{\url{https://help.luogu.com.cn/ula/luogu}} and Nowcoder,\footnote{\url{https://static.nowcoder.com/protocol/register.html}}\footnote{\url{https://static.nowcoder.com/protocol/privacy-policy.html}} ensuring that all experiments were conducted solely for academic purposes. Out of ethical and moral considerations, we will release the complete testing benchmark after conducting a comprehensive evaluation of the experimental code, data, and procedures, thereby facilitating academic exchange and technological advancement.

\bibliography{custom}

\clearpage

\appendix

\section{Problem Description} \label{sec:prob_d}

Figure \ref{fig:prob_desc} provides the problem descriptions we obtained for Codeforces problem \href{https://codeforces.com/problemset/problem/2043/D}{2043/D} as an example.

\section{Attribute Integration Details} \label{sec:integration}

Table \ref{tab:difficulty_levels} shows the complete difficulty classification of the problems. The ones from Luogu are originally in Chinese on the original website, and we have translated them into English for display. The classification is based on award tiers corresponding to ICPC, though it does not constitute a rigorous standard in practical applications.

Table \ref{tab:algorithm_categories} presents a classification of complete algorithm labels, with each category summarizing the complete algorithm labels from three websites. For Chinese labels, we have translated them into English.

\section{Model Details} \label{sec:model_d}

Table \ref{tab:models} presents the details of the models we evaluated, including the model link, number of parameters, type, and release date.

\section{Prompts} \label{sec:prompt}

We use the prompts of Figure \ref{fig:prompt} for generating model responses. The Chinese question descriptions are translated into English and then used accordingly.

\section{Error Category} \label{sec:error_c}

We classify errors on Codeforces and Nowcoder as follows: $``\text{COMPILATION\_ERROR}"$, $``\text{RUNTIME\_ERROR}"$, $``\text{IDLENESS\_LIMIT\_EXCEEDED}"$, $``\text{Execution error}"$, $``\text{Segmentation fault}"$, and $``\text{Floating point error}"$ are categorized as code errors. $``\text{WRONG\_ANSWER}"$, $``\text{TIME\_LIMIT\_EXCEEDED}"$, and $``\text{MEMORY\_LIMIT\_EXCEEDED}"$ are categorized as reasoning errors.

\section{Reasoning Complexity Hierarchy} \label{sec:rch}

Significant variations exist in the reasoning complexity of data structure and algorithm problems across different difficulty levels. For instance, in easy level problems, Math problems typically require single-step reasoning, demonstrating lower cognitive demands than Search problems. However, in medium and hard levels, Math problems frequently involve sophisticated data structures and algorithms, exhibiting substantially higher reasoning complexity than Search problems. Given the model's inferior performance on medium and hard levels problems, this study exclusively analyzed easy level problems.

At the easy difficulty level, dynamic programming (DP) problems generally necessitate the identification of state transition equations, which demands strong analytical capabilities and exceptional problem decomposition skills, thereby manifesting higher reasoning complexity. In contrast, data structure and graph-related problems, while requiring abstract thinking and modeling competencies, can often be resolved through template-based approaches at this level, resulting in marginally lower reasoning complexity compared to DP. Basic, String, and Math problems demonstrate relatively lower complexity in easy contexts. Our comprehensive evaluation established the following reasoning complexity hierarchy for east level problems: String < Basic < Math < Graph < Data Structures (DS) < Search < Dynamic Programming (DP).

\begin{figure*}[ht]
\centering
\begin{tcolorbox}[enhanced,size=small,colback=black!5!white,flip title={interior hidden},title={Problem description of Codeforces problem \href{https://codeforces.com/problemset/problem/2043/D}{2043/D}}]
\begin{lstlisting}[basicstyle=\ttfamily, captionpos=t, breaklines=true]
# Problem about GCD

## Problem Description
Given three integers $ l $ , $ r $ , and $ G $ , find two integers $ A $ and $ B $ ( $ l \le A \le B \le r $ ) such that their greatest common divisor (GCD) equals $ G $ and the distance $ |A - B| $ is maximized.
If there are multiple such pairs, choose the one where $ A $ is minimized. If no such pairs exist, output "-1 -1".

## Input Format
The first line contains a single integer $ t $ ( $ 1 \le t \le 10^3 $ ) - the number of test cases. Then, $ t $ test cases follow.
Each test case consists of a single line containing three integers $ l, r, G $ ( $ 1 \le l \le r \le 10^{18} $ ; $ 1 \le G \le 10^{18} $ ) - the range boundaries and the required GCD.

## Output Format
For each test case, output two integers $ A $ and $ B $ - the solution to the problem, or "-1 -1" if no such pair exists.

## Sample #1
### Sample Input #1
```
4
4 8 2
4 8 3
4 8 4
5 7 6
```

### Sample Output #1
```
4 6
-1 -1
4 8
6 6
```

## Limit
Time Limit
1.00s
Memory Limit
250.00MB
\end{lstlisting}
\end{tcolorbox}
\caption{Problem description.}
\label{fig:prob_desc}
\end{figure*}

\begin{table*}[ht]
    \centering
    \begin{tabular}{cccc}
        \toprule
        Platform & Easy & Medium & Hard \\
        \midrule
        Codeforces & $[800,1500]$ & $(1500,2400]$ & $(2400,3500)$ \\
        \midrule
        \multirow{2}{*}{Luogu} & Beginner, Basic- & Intermediate+/Advanced & Provincial/NOI- \\
        & Basic/Intermediate- & Advanced+/Provincial- & NOI/NOI+/CTSC \\
        \midrule
        Nowcoder & $[0,1500]$ & $(1500,2300]$ & $(2300,\infty)$ \\
        \midrule
        ICPC & Regional & Regional gold & \multirow{2}{*}{Winning the final} \\
        award tiers & bronze and silver & Eligibility for final & \\
        \bottomrule
    \end{tabular}
    \caption{Difficulty levels of different platforms.}
    \label{tab:difficulty_levels}
\end{table*}

\begin{table*}[ht]
    \centering
    \begin{tabular}{cc}
        \toprule
        Category & Algorithm tags \\
        \midrule
        \multirow{5}{*}{Basic} & 
        greedy, implementation, brute force, constructive algorithms, sortings, \\ & 
        two pointers, divide and conquer, bitmasks, simulation, construction,  \\ & 
        enumeration, recursion, two-pointer, thinking, violence, divide and conquer, \\ & 
        base conversion, Ad-hoc, classification discussion, bitwise operations, \\ & 
        randomization, discretization, prefix sums, differences \\  
        \midrule
        \multirow{2}{*}{Search} & 
        dfs and similar, meet-in-the-middle, binary search, ternary search, search, \\ & 
        breadth-first search BFS, depth-first search DFS, breadth-first search(BFS) \\  
        \midrule
        \multirow{2}{*}{String} & 
        strings, string suffix structures, expression parsing, trie trees,  \\ & 
        kmp and extended kmp, suffix arrays(SA), string hash \\  
        \midrule
        \multirow{2}{*}{DP} & 
        dynamic programming, dynamic programming DP, knapsack DP, tree DP,  \\ &
        dp, state compression enumeration, knapsack problem, interval dp \\  
        \midrule
        \multirow{5}{*}{DS} & 
        data structures, trees, dsu, hashing, STL, segment trees, scanlines, \\ & 
        tree theory, monotonic stacks, violent data structures, union-find,  \\ &
        queues, heaps, data structures, trees, stacks, binary indexed trees,  \\ &
        ST tables, doubling, chairman trees, hash, blocking, balanced trees,  \\ &
        tree of trees, RMQ, square root decomposition \\  
        \midrule
        \multirow{4}{*}{Graph} & 
        graphs, shortest paths, graph matchings, flows, 2-sat, graph theory, \\ &
        connected components, topological sorting, graph modeling, Tarjan, \\ &
        biconnected components, square trees, spanning trees, connectivity, \\ &
        shortest paths, network flows, tree decomposition \\  
        \midrule
        \multirow{8}{*}{Math} & 
        math, number theory, combinatorics, geometry, matrices, probabilities, \\ &
        fft, chinese remainder theorem, games, number theory, combinatorics, \\ &
        harmonic series, indeterminate equations, pigeonhole principle, \\ &
        linear algebra, computational geometry, probability expectation, \\ &
        prime factorization, gcd and exgcd, mobius inversion, linear basis,  \\ &
        inclusion-exclusion principle and pigeonhole principle, sieve method, \\ &
        quadrilateral inequality, mathematics, matrix multiplication, \\ &
        game theory, polynomials, schedules \\  
        \bottomrule
    \end{tabular}
    \caption{Algorithm categories and their topics.}
    \label{tab:algorithm_categories}
\end{table*}

\begin{table*}[ht]
    \centering
    \begin{tabular}{cccccc}
        \toprule
        Short Name & Model & Parameter & Type & Data \\
        \midrule
        QwQ & \href{https://huggingface.co/Qwen/QwQ-32B-Preview}{Qwen/QwQ-32B-Preview} & 32B & Reasoning & 2024/11 \\
        DS-V3 & \href{https://huggingface.co/deepseek-ai/DeepSeek-V3}{deepseek-ai/DeepSeek-V3} & 37/671B & Instruction & 2024/12 \\
        Qwen2.5-72B & \href{https://huggingface.co/Qwen/Qwen2.5-72B-Instruct}{Qwen/Qwen2.5-72B-Instruct} & 72B & Instruction & 2024/09 \\
        Mistral-Large & \href{https://huggingface.co/mistralai/Mistral-Large-Instruct-2411}{mistralai/Mistral-Large-Instruct-2411} & 123B & Instruction & 2024/11 \\
        Qwen2.5-Coder & \href{https://huggingface.co/Qwen/Qwen2.5-Coder-32B-Instruct}{Qwen/Qwen2.5-Coder-32B-Instruct} & 32B & Code & 2024/11 \\
        Llama-3.1-70B & \href{https://huggingface.co/meta-llama/Llama-3.1-70B-Instruct}{meta-llama/Llama-3.1-70B-Instruct} & 70B & Instruction & 2024/07 \\
        Codestral-v0.1 & \href{https://huggingface.co/mistralai/Codestral-22B-v0.1}{mistralai/Codestral-22B-v0.1} & 22B & Code & 2024/05 \\
        Skywork-o1 & \href{https://huggingface.co/Skywork/Skywork-o1-Open-Llama-3.1-8B}{Skywork/Skywork-o1-Open-Llama-3.1-8B} & 8B & Reasoning & 2024/11 \\
        Mixtral-8x22B & \href{https://huggingface.co/mistralai/Mixtral-8x22B-Instruct-v0.1}{mistralai/Mixtral-8x22B-Instruct-v0.1} & 22/176B & Instruction & 2024/04 \\
        \bottomrule
    \end{tabular}
    \caption{Model details.}
    \label{tab:models}
\end{table*}

\begin{figure*}[ht]
\centering
\begin{tcolorbox}[enhanced,size=small,colback=black!5!white,flip title={interior hidden}]
\begin{lstlisting}[basicstyle=\ttfamily, label={}, captionpos=t, breaklines=true]
messages = [
    {
        "role": "system",
        "content": "You are a highly skilled competitive programming expert, adept at analyzing complex problems and designing efficient solutions. Your task is to solve the following programming challenge. Always maintain clear logic and think step by step."
    },
    {
        "role": "user",
        "content": """
            Your task is to carefully read the following problem description, analyze the problem step by step, and clearly explain your thought process. Finally, write the solution in {lang} and ensure the code is correct and readable. Wrap your code in the following format:
            
            ```{lang_type}
            {lang} solution code
            ```
            
            Here is the problem description:
            
            {description}
            '''
        """
    }
]
\end{lstlisting}
\end{tcolorbox}
\caption{Prompts for generating model responses.}
\label{fig:prompt}
\end{figure*}

\end{document}